# Inter-Feature-Map Differential Coding of Surveillance Video

Kei Iino
*Graduate School of FSE*
*Waseda University*
Tokyo, Japan
iinokei@akane.waseda.jp

Miho Takahashi
*Graduate School of FSE*
*Waseda University*
Tokyo, Japan
miho.takahashi@akane.waseda.jp

Hiroshi Watanabe
*Graduate School of FSE*
*Waseda University*
Tokyo, Japan
hiroshi.watanabe@waseda.jp

Ichiro Morinaga
*NTT Software Innovation Center*
Tokyo, Japan
ichiro.morinaga.tv@hco.ntt.co.jp

Shohei Enomoto
*NTT Software Innovation Center*
Tokyo, Japan
shohei.enomoto.ab@hco.ntt.co.jp

Xu Shi
*NTT Software Innovation Center*
Tokyo, Japan
xu.shi.ca@hco.ntt.co.jp

Akira Sakamoto
*NTT Software Innovation Center*
Tokyo, Japan
akira.sakamoto.ax@hco.ntt.co.jp

Takeharu Eda
*NTT Software Innovation Center*
Tokyo, Japan
takeharu@acm.org

*Abstract*— In Collaborative Intelligence, a deep neural network (DNN) is partitioned and deployed at the edge and the cloud for bandwidth saving and system optimization. When a model input is an image, it has been confirmed that the intermediate feature map, the output from the edge, can be smaller than the input data size. However, its effectiveness has not been reported when the input is a video. In this study, we propose a method to compress the feature map of surveillance videos by applying inter-feature-map differential coding (IFMDC). IFMDC shows a compression ratio comparable to, or better than, HEVC to the input video in the case of small accuracy reduction. Our method is especially effective for videos that are sensitive to image quality degradation when HEVC is applied.

## I. Introduction

These days, mobile and IoT devices [1], called edge devices, increasingly depend on artificial intelligence (AI) engines for advanced applications such as autonomous cars, smart cities, and personal digital assistants [2]. Deep neural networks (DNNs) are commonly used as AI engines. While such DNN models are highly accurate, they require a lot of computational resources. Therefore, the most common approach is to place the model in the cloud and it process sensor data (images, texts, audio, etc.) compressed and uploaded by edge devices. This is called the Cloud-AI approach (Fig. 1 upper). Recently, edge devices have been equipped with small GPUs to enable some DNN models to run at the edge. This is called the Edge-AI approach (Fig. 1 bottom).

In recent years, another approach called Collaborative Intelligence [3] (Fig. 2) has been proposed for bandwidth saving and system optimization. This is a possibility between the Cloud-AI and Edge-AI. In this approach, a DNN model is partitioned into two parts, the front end deployed at the edge, and the back end deployed at the cloud. The front end consists of an input layer and several subsequent layers. The back end consists of the remaining layers. To perform AI tasks, the intermediate feature maps, outputs from the edge side, are uploaded to the cloud for the rest of the computation.

When a model input is an image, it has been reported that the feature map from the edge can be compressed smaller than the input data size [4,5,6]. One important use case for DNN at the edge is the analysis of surveillance video, but a feature compression method that takes advantage of video characteristics has not been reported yet, to our knowledge. Feature maps of a video can be compressed by removing the temporal correlations as with video coding. In this study, we adopt inter-feature-map differential coding (IFMDC) to compress the feature map of a surveillance video. IFMDC is a simple and lightweight method to take the residuals from the adjacent frame and quantize it as with DPCM [7]. We confirm its effectiveness in the object detection task with some baselines.

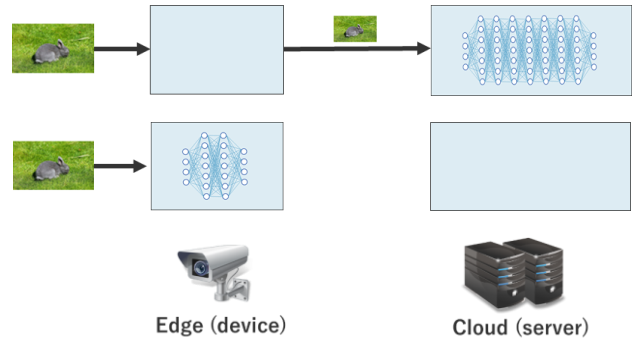

Fig. 1. Upper: cloud-only, bottom: edge-only

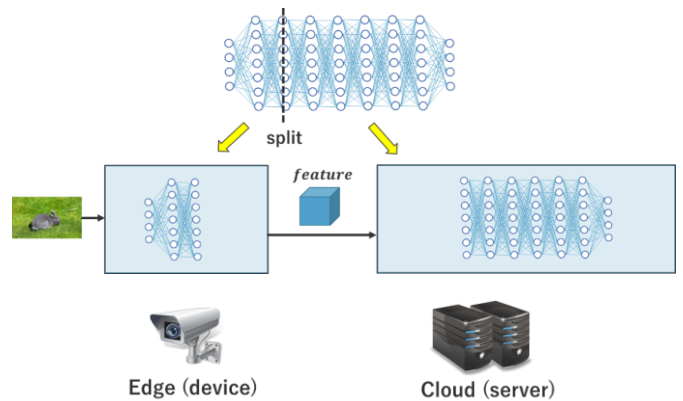

Fig. 2. Collaborative Intelligence

## II. Related Work

### A. Compression method using existing codec

Choi et al. [4] rearranged the feature map (H×W×C) to produce gray-scale images and interpreted them as a video that is composed of more than 1 frame (Fig. 3). They tested some rearrangement methods, such as tiling, where channels of the feature map are placed in the image as a tile, followed by another tile, and so on, and quilting, where neighboring samples come from different channels. They reported tiling on the single image was the best way and encoded it by HEVC [8] intra-mode after quantization. By re-training the model considering quantization, the data size after this compression was below the JPEG input up to 70%.

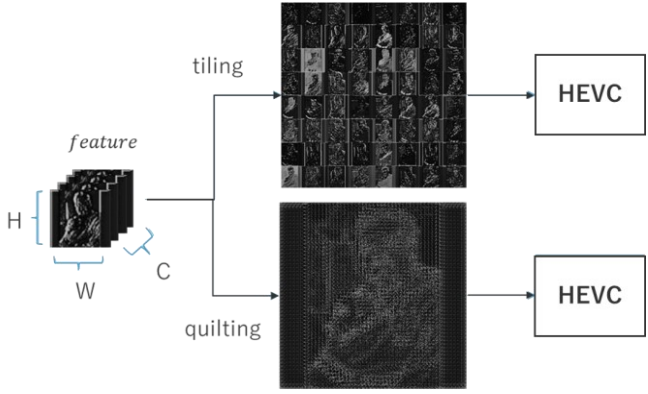

Fig. 3. Compression method using existing code (HEVC) to the rearranged feature map (tiling and quilting) [4].

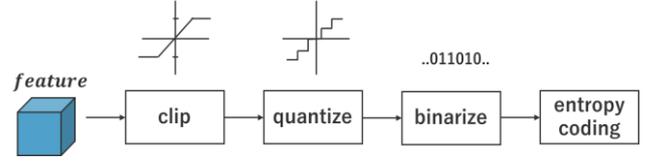

Fig. 4. Lightweight compression by clipping and quantization [5].

### B. Lightweight compression by clipping and quantization

Cohen et al. applied the sequence of operations (clipping, quantization, binarization, and entropy coding) to the feature map, as shown in Fig. 4. [5]. In the experiment, they tried some sets of a clip range and a quantization level for the feature map from the activation layer. Then, they succeeded in compressing it more than the method in [4], from 32-bit floating point down to 0.6 to 0.8 bits, while keeping the loss in accuracy to less than 1%. This method is lightweight and does not need retraining, so it seems to be suitable for the edge.

### III. PROPOSED METHOD

We adopt inter-feature-map differential coding (IFMDC) to remove the temporal correlation of the feature maps. IFMDC is the same approach as differential pulse-code modulation (DPCM). These methods are simple and lightweight compression methods to take the residuals from the adjacent frame and quantize it. It is known that quantization of the feature maps doesn't have great effects on the task accuracy [4] and the combination of it with clipping is effective [5]. In DNNs, the spatial size (H x W) of the feature maps gets smaller at deeper layers due to the convolution and pooling operation [9]. Because of this fact, the feature map of a frame is almost the same as that of the adjacent frame. In addition, a motion vector between frames of surveillance video tends to be small because the camera is fixed. Therefore, IFMDC, which is computationally small, is considered to work effectively for surveillance video.

IFMDC consists of the following steps, *1) ~ 3)*. The block diagram of it is shown in Fig. 5 ($n$ : frame number, $T$ : periodic time or GOP).

*1)* As a preprocessing step, the feature map is clipped into the certain value range:

$$if\ feature(n,p) \geq clip_{max}:$$
$$feature(n,p) = clip_{max}, \quad (1)$$

$$if\ feature(n,p) \leq clip_{min}:$$
$$feature(n,p) = clip_{min}, \quad (2)$$

where $feature(n,p)$ is the p-th pixel of the feature map of the n-th frame and $clip_{max}/clip_{min}$ is the clipping value.

*2)* If it is the feature map of non-key frame ($n \bmod T \neq 0$) the residual feature map is created by taking difference between adjacent feature maps:

$$if\ n \bmod T \neq 0:$$
$$residual(n) = feature(n) - feature(n-1)', \quad (3)$$

where $feature(n)'$ is the reconstructed feature map.

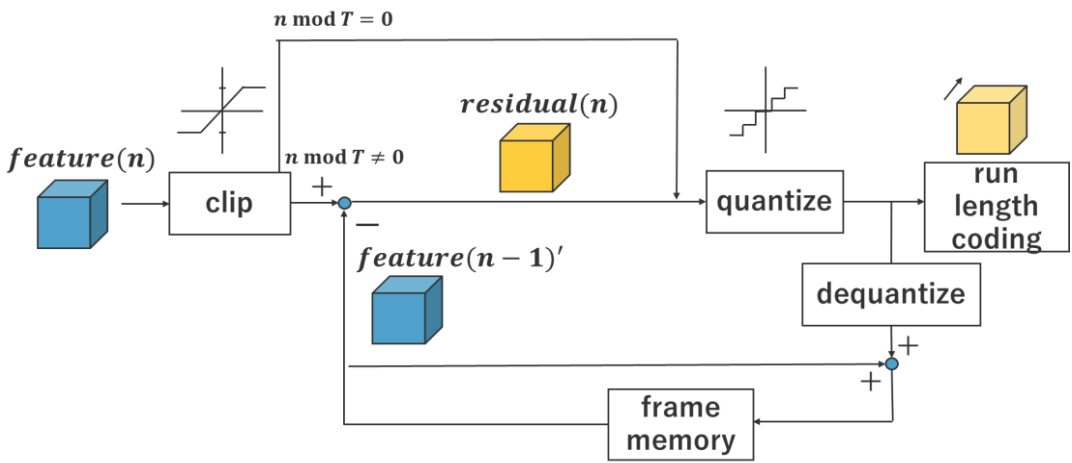

Fig 3. Block diagram of IFMDC

*3)* Then, these maps are quantized and compressed using run-length encoding. Especially, we use N-run-length encoding because the values of the residual are almost the certain symbol, N, after the quantization (Fig. 6).

$$A = clip_{max} - clip_{min}, \quad (4)$$

$$\widetilde{residual}(n) = round\left(\frac{residual(n) - A}{2A} \cdot (k-1)\right), \quad (5)$$

where $A$ is the clip range and $k$ is the quantization level.

## IV. EXPERIMENT

### A. Evaluation

We evaluate the performance of IFMDC and three baselines in the object detection task. In the experiment, each method is evaluated in terms of rate-accuracy tradeoff. As the detection model, we use YOLOv3 [10] and split it at the 12th layer of the model. The input image size is set to (C, H, W) = (3, 416, 416) and the size of the feature map from the split layer is (256, 52, 52).

### B. Baselines and Setting

As the baselines, we use HEVC compression of the input video (HEVC-video) and of the feature maps rearranged by tiling and quilting (HEVC-tiling, HEVC-quilting) [4]. For each HEVC compression, we use FFmpeg [11] with GOP=15, no b-frames, and coding-tree-unit (CTU) size is set to 16 for HEVC-tiling/quilting. The data size of each method is calculated only for p-frames. The compression ratio is controlled by the QP-parameter in the baselines, and the quantization level, $k$, in IFMDC. Since the clipping and quantization in the middle layer contribute to the improvement of accuracy in this experiment, we apply the same process to the baselines, not only to IFMDC, for a fair comparison.

### C. Dataset

The videos used in the experiment are 27 sequences from MOTSynth [12], which is a full-HD synthetic video dataset for pedestrian detection and tracking.

For each video, we show the influence of the HEVC compression on the detection accuracy: the difficulty of encoding with HEVC (Fig. 7). Each line corresponds to one video. The horizontal axis shows the compression ratio (6), and the vertical axis shows the rate of AP loss on a logarithmic scale (7):

$$Compression\ ratio = \frac{I_{after}}{I_{before}}, \quad (6)$$

$$AP\ loss\ rate = 10\ log_{10}\frac{AP_{before} - AP_{after}}{AP_{before}}, \quad (7)$$

where $I$ represents the data size, and the subscripts $before$ and $after$ represent before and after the compression, respectively. In this graph, the videos on the upper right are considered more sensitive to image quality degradation: they are cases where the detection accuracy easily drops with image quality loss by HEVC compression.

### D. Result

The averaged results for all videos are shown in Fig. 8. The vertical axis shows the average precision (AP) and the horizontal axis shows the bits per pixel (BPP). As can be seen from this figure, IFMDC has an advantage over all baselines when the AP loss is under 2%.

## V. DISCUSSION

To analyze the result, we assign blue and red labels to the videos in Fig. 9. Blue labels are assigned to the videos where IFMDC outperforms HEVC-video, and red labels are assigned to the remaining videos. As can be seen from this figure, IFMDC is particularly effective for videos that are sensitive to image quality degradation, for example, videos containing very small objects or objects that easily assimilate into the background (Fig. 10 right). This is considered due to the difference between video compression and feature compression. If compression is performed to the video at the input stage, information like edge, important for recognition, may be lost. On the other hand, if compression is performed to the feature map at the intermediate layer, such information loss is considered to be relatively small. This consideration is reinforced by the fact that HEVC-tiling/quilting tend to outperform HEVC-video in the blue labels (Fig. 9) as well as IFMDC.

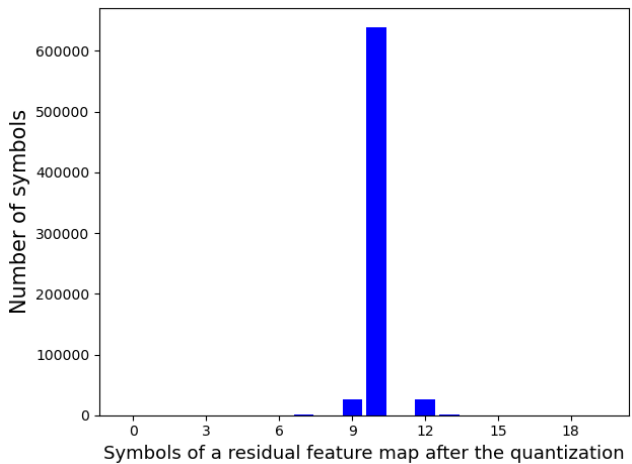

Fig 6. Distribution of the residual map symbols after the quantization.

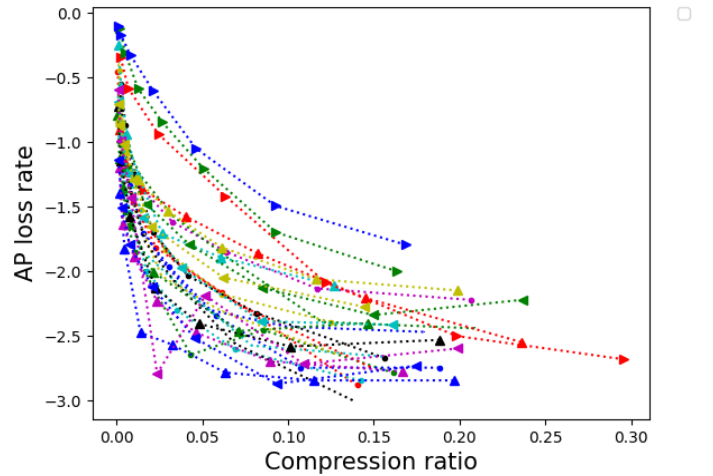

Fig 7. The influence of the HEVC compression to the detection accuracy.

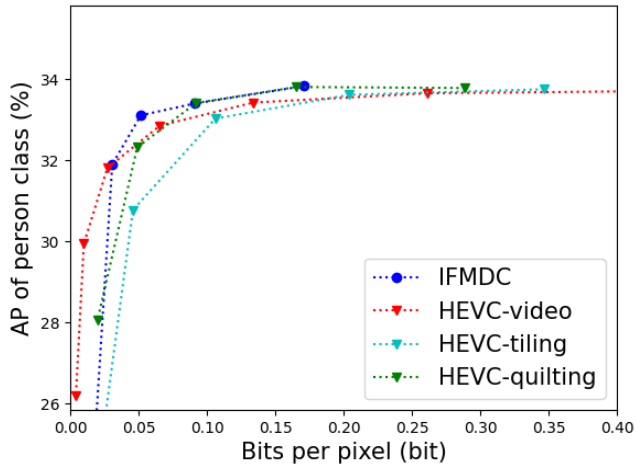

Fig 8. Object-detection AP of YOLOv3 and BPP of each method.

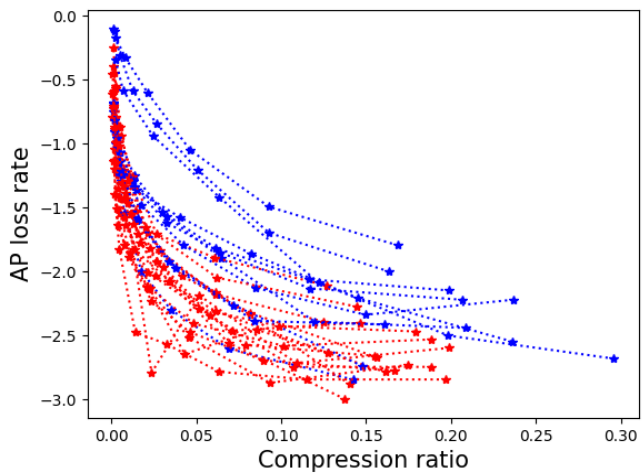

Fig 9. Comparision of IFMDC and HEVC-video by labeling the videos in Fig. 7. Blue: IFMDC outperforms HEVC-video, Red: HEVC-video outperforms IFMDC.

## VI. Conclusion

In this paper, we proposed IFMDC which is a video feature compression method for collaborative intelligence. IFMDC takes the same approach as DPCM. We evaluated the method with three baselines in the object detection task of surveillance videos. Even though it is very simple and lightweight, IFMDC is comparable to, or better than, HEVC compression of the input video in our experiment. Analysis of the results of each video showed that our method is especially effective for videos that are sensitive to image quality degradation: videos containing very small objects or objects that easily assimilate into the background. As a future task, we are planning to support larger objects and videos with more motion.

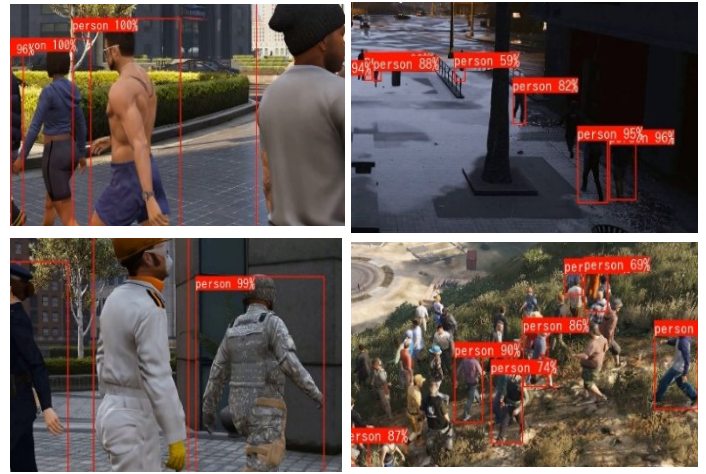

Fig 10. Cropped images of the output video frames. Left: from the videos labeled red in Fig. 9, Right: from the videos labeled blue in Fig. 9.